\documentclass[letterpaper]{article} 
\usepackage[preprint]{aaai2027}  
\usepackage[hyphens]{url}  
\usepackage{graphicx} 
\graphicspath{{./}{figures/}{Figures/}}
\usepackage{natbib}  
\usepackage{caption} 
\usepackage{algorithm}
\usepackage{algpseudocode}
\renewcommand{\Call}[2]{\textproc{#1}(#2)}

\usepackage{booktabs}
\usepackage{tabularx}

\usepackage{amsmath,amssymb}
\usepackage{xcolor}
\usepackage{tikz}
\usetikzlibrary{arrows.meta,positioning,fit,calc,shapes.geometric,patterns}

\title{GeoNest: Learning to Select Failure-Aware Neighborhoods for the Irregular Knapsack Problem in a Circular Container}

\author{
    Zhongman Du\textsuperscript{\rm 1},
    Huiming Zhang\textsuperscript{\rm 1},
    Linlin Yang\textsuperscript{\rm 2},
    Sheng Xu\textsuperscript{\rm 2},
    Baochang Zhang\textsuperscript{\rm 1,3}
}

\affiliations{
    \textsuperscript{\rm 1}Beihang University, Beijing, China\\
    \textsuperscript{\rm 2}Communication University of China, Beijing, China\\
    \textsuperscript{\rm 3}Hangzhou Innovation Institute of Beihang University, Hangzhou, China
}

\newcommand{\method}{\textsc{GeoNest}}
\newcommand{\bench}{\textsc{CircleNest-Bench}}
\newcommand{\solver}{\textsc{PackingSolver}}

\begin{document}

\maketitle

\begin{abstract}
The two-dimensional irregular knapsack problem in a fixed circular container is an important combinatorial optimization problem for maximizing material utilization in manufacturing. Conventional geometric packing solvers can produce tightly packed layouts, yet they often partition the residual space into isolated small pockets that cannot fit valuable unplaced polygons. To overcome this late-stage packing bottleneck, we propose a failure-aware large neighborhood search framework named GeoNest, driven by a graph policy trained via reinforcement learning. Specifically, we first construct neighborhoods by pairing failed target polygons with residual pockets. We then use explanatory poses to identify the placed polygons that block candidate insertions. These diagnosed blocking relations define bounded, fixed-item repair subproblems for the underlying geometric solver. Finally, the graph policy selects the most promising subproblem for execution. For evaluation, we introduce CircleNest-Bench, a benchmark comprising 2,391 load-controlled instances from four contour sources, including a held-out industrial CAD source. Experimental results demonstrate that, under the same total time budget, GeoNest improves mean utilization over a state-of-the-art standalone packing solver by about 0.9\% on average across the three main test sets and by about 0.6\% on the held-out industrial set.
\end{abstract}

\section{Introduction}

Two-dimensional irregular packing is a classical NP-hard problem with broad applications in material cutting, industrial nesting, and texture atlas generation \citep{lee2008nesting,noll2011atlas}. Most studies consider strip or rectangular containers, whose straight boundaries and principal directions provide natural references for piece placement \citep{leao2020review}. In contrast, substrates such as semiconductor wafers and optical lens blanks are often circular or nearly circular and lack corner structure. Prior circular-container studies mainly consider regular objects \citep{huang2006circles,lopez2018rectangles}. To our knowledge, this is the first study of irregular polygon knapsack
packing in a fixed circular container. We call this problem the
Two-Dimensional Circular-Container Irregular Knapsack Problem (2D-CIKP). Given an overloaded candidate set, 2D-CIKP selects a subset and determines translations and continuous rotations to maximize packed area subject to containment and non-overlap.

Traditional methods combine explicit geometric computation with constructive heuristics or metaheuristic search, but early placement decisions can lock the search into unfavorable residual-space structures \citep{gomes2006hybrid,sato2019raster}. Recent generative approaches jointly refine multiple piece poses \citep{xue2023gfpack,xue2025gfpackpp}, while reinforcement learning methods optimize packing sequences or select predefined low-level operators \citep{fang2023hybridrl,wang2026hhql}. These approaches do not turn the interaction between an unpacked target and the residual geometry into a targeted repair decision.

\begin{figure}[t]
    \centering
    \includegraphics[width=\columnwidth]{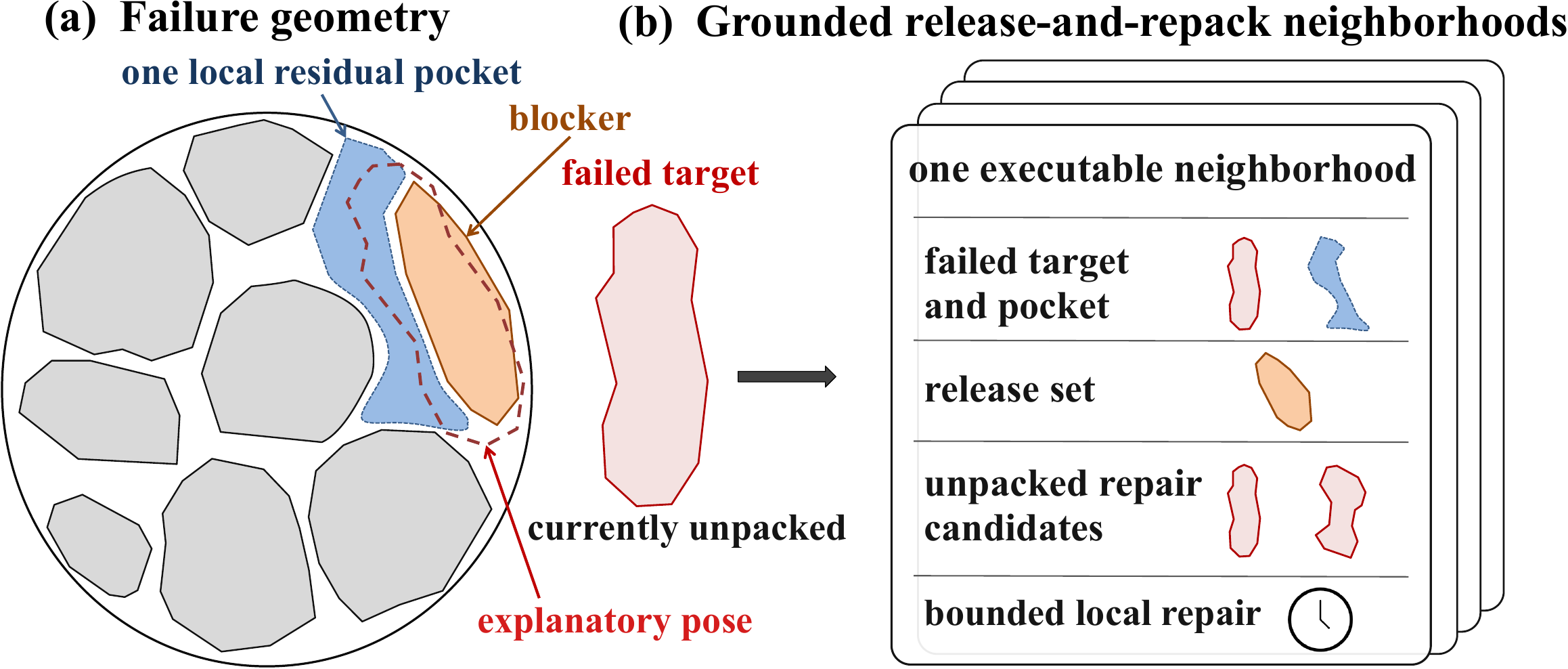}
    \caption{Failure geometry and grounded repair in 2D-CIKP. A target--pocket pair and its explanatory pose identify blockers, which define executable release-and-repack neighborhoods.}
    \label{fig:failure}
\end{figure}

This limitation is most acute in the late stages of search. Geometry-based solvers quickly construct high-quality feasible layouts, but further improvement becomes difficult as utilization rises and residual free space fragments into pockets with blocked access or incompatible shapes. A valuable polygon can then remain unpacked even when the total residual area exceeds its area, while generic perturbations preserve the same bottleneck. As illustrated in Fig.~\ref{fig:failure}, pairing a failed target with a residual pocket exposes an explanatory pose and its blockers, turning failure geometry into concrete release-and-repack neighborhoods. The central late-stage decision is therefore which neighborhood to repair next.

\begin{figure*}[t!]
\centering
\includegraphics[width=0.95\linewidth]{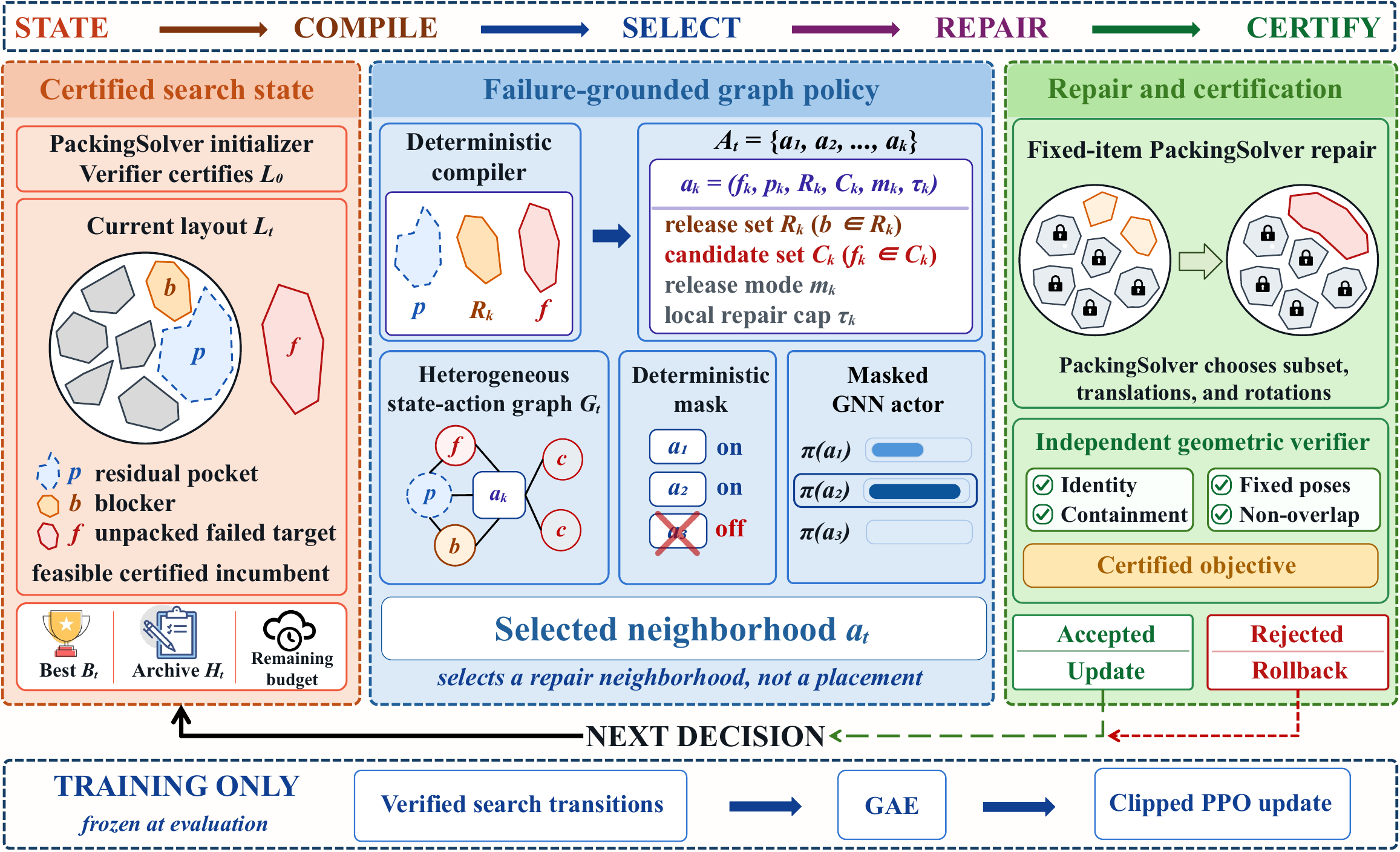}
\caption{Overview of one \method{} decision. An unpacked target is paired with a residual pocket, and an explanatory pose identifies candidate blockers. The resulting fixed-item repair actions are encoded in a graph and scored by the policy. \solver{} then executes the selected action under its repair-time cap.}
\label{fig:overview}
\end{figure*}

We therefore propose \method{}, a failure-aware, learning-guided large-neighborhood search framework that augments geometry-based solvers, as illustrated in Fig.~\ref{fig:overview}. Following learning-guided neighborhood search \citep{wu2021lns,reijnen2024dralns}, \method{} does not predict complete layouts or continuous poses. Starting from a feasible layout produced by bounded geometric search, it pairs unpacked targets with compatible residual pockets, constructs explanatory poses, and uses the resulting blockers to compile fixed-item repair neighborhoods at multiple scales. Each executable neighborhood forms an action defined by its release set, admitted unpacked items, scale, and repair-time cap.

The value of an action depends on the current target--pocket--blocker configuration. Each repair decision updates future eligible actions and the remaining budget and may change the residual geometry, so neighborhood selection is sequential. Targets, pockets, blockers, and actions are encoded in a heterogeneous graph, where a masked GNN policy trained with PPO ranks eligible actions. PackingSolver \citep{packingsolver} executes the selected repair with all other pieces fixed, and a common geometric verifier certifies the result.

For evaluation, we introduce \bench{}, comprising 2,391 load-controlled circular-container instances from four contour sources. It covers in-distribution testing, size extrapolation, overload stress testing, cross-source transfer, and held-out industrial CAD. All methods use identical instances and matched time budgets, and only verifier-certified layouts contribute utilization. \method{} improves certified utilization over the strongest geometry-based baseline by 0.6\% to 1.0\%. Ablations isolate the contributions of failure-aware neighborhood construction and learned neighborhood selection.
The main contributions are as follows.
\begin{itemize}

\item We formalize 2D-CIKP and introduce \bench{}, a benchmark of 2,391 load-controlled instances from four contour sources.

\item We introduce a failure-aware neighborhood construction mechanism that uses target--pocket pairs and explanatory poses to diagnose blockers and compile fixed-item repair neighborhoods.

\item We propose \method{}, which casts failure recovery as sequential selection among repair-neighborhood actions and uses a masked GNN policy trained with PPO to rank them. Experiments demonstrate improved certified utilization and generalization across instance sizes and contour sources.

\end{itemize}

\section{Related Work}

\subsection{Irregular Packing and Fixed-Container Problems}

Irregular packing relies on explicit geometric models of containment, non-overlap, and candidate placement. No-fit polygons encode pairwise relative-placement constraints, while raster models and mixed-integer formulations provide alternative mechanisms for geometric feasibility \citep{bennell2008survey,burke2007nfp,umetani2022raster,lastra2024vertical}. Large neighborhood search and adaptive large neighborhood search provide general frameworks for improving incumbent solutions \citep{pisinger2010lns,ropke2006alns}, and recent work applies LNS to irregular bin packing through large-scale perturbations of cross-bin polygon assignments \citep{wang2026lns2dibpp}. Open-source heuristics such as \textsc{Sparrow} improve reproducibility \citep{gardeyn2025sparrow}.

Representative circular-container studies consider packing regular objects, including unequal circles and rectangles, in a fixed circular container \citep{lopez2016circles,bouzid2020rectangles}. The closest prior formulation we are aware of is the \textsc{Maximum Polygon Packing} problem from the CG:SHOP 2024 Challenge \citep{fekete2024mpp}, addressed by representative SoCG approaches \citep{atak2024mpp,dafonseca2024shadoks}. It considers profit-maximizing subset selection in a fixed convex polygonal container but permits translations only. 2D-CIKP instead considers a fixed circular container with continuous rotations, making radial boundary effects and the orientation-dependent accessibility of residual pockets central to subsequent insertions.

\subsection{Learning-Guided Packing and Neighborhood Search}

Learning-based irregular packing methods act during layout construction or refinement. Reinforcement learning has been used to optimize piece sequences \citep{fang2023hybridrl}, select low-level operators for sequence and orientation optimization \citep{wang2026hhql}, and learn rotational placement within hybrid nesting pipelines \citep{abdou2026nest}. Yang et al.\ select and group irregular UV patches and predict their relative poses, whereas GFPack and GFPack++ learn gradient fields for joint layout refinement \citep{yang2023learn2pack,xue2023gfpack,xue2025gfpackpp}. In contrast, the learned policy in \method{} operates after bounded geometric search and uses unpacked items as failure targets without predicting placement coordinates or rotations.

A complementary line preserves a conventional solver and learns to control large neighborhood search. For integer programs, learned policies select neighborhoods or variables for solver-based reoptimization \citep{song2020lns,wu2021lns}. Neural large neighborhood search learns repair heuristics, while learned ALNS methods control operators, parameters, or acceptance decisions \citep{hottung2022neural,reijnen2024dralns,johnn2024grlos}. \method{} follows this solver-control principle but uses a different action interface. Instead of selecting a global operator or a generic subset of solver variables, it ranks repair actions constructed from the current failed target and its local geometric context.

\section{Problem Formulation and Benchmark}
\label{sec:problem}

\paragraph{Task formulation.} An instance of 2D-CIKP consists of a fixed circular container and $n$ candidate polygonal items. Let $[n]=\{1,\ldots,n\}$ and let $P_i\subset\mathbb{R}^2$ be a compact simple polygonal region with area $a_i=\operatorname{area}(P_i)>0$. The container is the fixed disk $D_R=\{x\in\mathbb{R}^2:\lVert x\rVert_2\le R\}$ with radius $R>0$. A selected item $i$ is assigned a pose $q_i=(\mathbf{u}_i,\theta_i)\in\mathbb{R}^2\times[0,2\pi)$, where $\mathbf{u}_i$ is a continuous translation vector and $\theta_i$ may take any in-plane angle over the full $360^\circ$ range. Its transformed region is $X_i(q_i)=Q(\theta_i)P_i+\mathbf{u}_i$, where $Q(\theta_i)$ is the planar rotation matrix. Reflections are not allowed. A layout is $L=\{(z_i,q_i)\}_{i=1}^n$, where $z_i\in\{0,1\}$ indicates whether item $i$ is selected and $q_i$ is immaterial when $z_i=0$. The layout is feasible if the following conditions hold for every $i\in[n]$ and every pair $i<j$:
\[
\begin{aligned}
z_i=1 &\ \Longrightarrow\ X_i(q_i)\subseteq D_R,\\
z_i=z_j=1 &\ \Longrightarrow\ \operatorname{int}X_i(q_i)\cap\operatorname{int}X_j(q_j)=\varnothing.
\end{aligned}
\]

Boundary contact is allowed. For a feasible layout, define its disk utilization as
\[
U(L)=\frac{\sum_{i=1}^n z_i a_i}{\pi R^2}.
\]

Let $\mathcal{L}_{\mathrm{feas}}$ denote the set of feasible layouts. We solve
\[
L^\star\in\operatorname*{argmax}_{L\in\mathcal{L}_{\mathrm{feas}}}U(L).
\]
Since $R$ is fixed for each instance, maximizing $U(L)$ is equivalent to maximizing the total packed area. The offered load is
\begin{equation}
\lambda=\frac{\sum_{i=1}^n a_i}{\pi R^2},\qquad R=\sqrt{\frac{\sum_{i=1}^n a_i}{\pi\lambda}}.
\label{eq:offered-load}
\end{equation}
CircleNest-Bench uses $\lambda>1$, so the total candidate area exceeds the disk area and subset selection is necessary.

\paragraph{CircleNest-Bench.} Public irregular-packing collections provide diverse source contours for constructing controlled circular-container knapsack instances across size, load, and source shifts. To provide a reproducible evaluation basis, CircleNest-Bench contains 2,391 load-controlled 2D-CIKP instances generated from ESICUP shapes, controlled synthetic shapes, the Gardeyn nesting instances, and a held-out industrial geometry source \cite{esicup_datasets,gardeyn2025sparrow}. The benchmark converts the source contours into new circular-container knapsack instances and sets the disk radius from Eq.~\eqref{eq:offered-load} after forming the candidate set. Tab.~\ref{tab:benchmark-splits} reports the complete split statistics.

\begin{table}[htbp]
\centering
{\small
\begin{tabular*}{\columnwidth}{@{\extracolsep{\fill}}lrrrrr@{}}
\toprule
Split & E & S & G & I & Total \\
\midrule
Train & 720 & 720 & 360 & -- & 1,800 \\
Validation & 90 & 90 & 54 & -- & 234 \\
Main & 90 & 90 & 54 & -- & 234 \\
Industrial & -- & -- & -- & 14 & 14 \\
Size & 40 & 40 & 4 & -- & 84 \\
Stress & 10 & 10 & 5 & -- & 25 \\
\midrule
Total & 950 & 950 & 477 & 14 & 2,391 \\
\bottomrule
\end{tabular*}
}
\caption{CircleNest-Bench partitions. E, S, G, and I denote the ESICUP, Synthetic, Gardeyn, and Industrial sources, respectively. The Industrial source contains held-out CAD-derived contours.}
\label{tab:benchmark-splits}
\end{table}

\paragraph{Splits and evaluation.}
The partitions are disjoint at the base candidate-item-set level. All load variants derived from the same base set share identical items, differ only in disk radius, and remain in the same split. Primary evaluation uses ESICUP Main, Synthetic Main, and Gardeyn Main and reports the unweighted macro average of their source-level means. The Main partitions span $n\in\{48,64,96\}$ and offered loads $\lambda\in\{1.2,1.5,2.0\}$. The Size split evaluates extrapolation to $n=128$ at $\lambda\in\{1.5,2.0\}$, while Stress fixes $n=64$ and tests $\lambda=2.5$. The held-out size $n=128$ and load $\lambda=2.5$ do not appear in training or validation. Industrial contains 14 test-only CAD instances with $n=100$ and $\lambda=1.5$. Its source contours do not occur in training or validation, and the Industrial split is excluded from the three-source Main average.

All methods submit layouts to a common verifier that reconstructs every transformed polygon from the immutable instance geometry, checks item identities, disk containment, and pairwise non-overlap, and recomputes $U(L)$. The primary metric is verifier-certified utilization, and invalid layouts receive zero. The benchmark release includes the instance files, split manifests, generation scripts, and verifier. Additional details on source mappings, instance construction, split auditing, and verification are provided in the supplementary material.

\section{Method}
\label{sec:method}

Given a feasible initial layout $L_0$ and a takeover budget, \method{} treats items left unpacked by bounded search as failure targets and pairs them with residual pockets to construct fixed-item repair subproblems. A graph policy ranks the corresponding solver calls and allocates the next portion of the budget to the selected subproblem. Fig.~\ref{fig:overview} illustrates the process.

\subsection{Failure-Grounded Neighborhoods}

At decision $t$, the controller state is
\[
s_t=(L_t,B_t,\mathcal{H}_t,\mathcal{F}_t,\mathcal{P}_t,\zeta_t),
\]
where $L_t$ and $B_t$ are the current and certified best layouts, respectively, $\mathcal{H}_t$ stores verifier-valid, geometrically distinct archived layouts, $\mathcal{F}_t$ orders items left unpacked by the preceding bounded search, $\mathcal{P}_t$ contains residual pockets, and $\zeta_t$ records the remaining budget, stagnation, and retry state.

Let $\mathcal{I}(L_t)$ denote the identities of items placed in $L_t$. Using a deterministic polygonal disk proxy $\widehat D_R$, we define the residual region as
\[
\widehat\Omega_t=\widehat D_R\setminus\bigcup_{i\in\mathcal{I}(L_t)}X_i(q_i).
\]
The nonempty connected components obtained by intersecting $\widehat\Omega_t$ with fixed multiscale radial and angular windows form $\mathcal{P}_t$. For a target $f$ and pocket $p$, the compatibility score $\rho(f,p)\in[0,1]$ combines their area ratio with the best axis-aligned bounding-box fit under the original and $90^\circ$-swapped spans and ranks target--pocket pairs for geometric testing.

For each retained pair $(f,p)$, the generator constructs finite orientation and translation-anchor sets and produces
\[
\mathcal Q_0(f,p)=
\{q=(\mathbf u,\theta)\mid
\theta\in\Theta(f,p),\
\mathbf u\in\mathcal T(f,p,\theta)\}.
\]
A pose is retained if the transformed target lies inside the disk and at least a fraction $\eta\in(0,1]$ of its area overlaps $p$. It may overlap incumbent items, whose identities define the pose-induced blockers
\[
\mathcal{B}_t(f,q)=
\left\{
i\in\mathcal{I}(L_t)
\ \middle|\
\operatorname{area}\!\left(X_f(q)\cap X_i(q_i)\right)>\varepsilon_{\mathrm{blk}}
\right\},
\]
where $\varepsilon_{\mathrm{blk}}>0$ is the blocker-overlap threshold. Among the retained poses, the explanatory pose $q^\star$ maximizes a fixed pose score that rewards pocket coverage and penalizes blocked area, blocker count, and radial displacement. Its blockers form the core release set, while the repair solver determines the final poses. Exact formulations of $\rho(f,p)$ and $q^\star$ are provided in the supplementary material.

Each executable failure-specific repair neighborhood defines a grounded action of the form
\[
a=(f_a,p_a,\mathcal{R}_a,C_a,m_a,\tau_a),
\]
where $f_a$ and $p_a$ are its target and pocket, $\mathcal{R}_a$ is the release set, $C_a\ni f_a$ is a bounded set of admitted unpacked items, and $\tau_a$ is the repair-time cap. The mode $m_a\in\{\textsc{core},\textsc{local},\textsc{connected},\textsc{deep-connected}\}$ determines $\mathcal{R}_a$. \textsc{core} releases only the pose-induced blockers, \textsc{local} also releases nearby placed items, and the connected modes expand through the placed-item connectivity graph at increasing depth. Additional unpacked items are admitted only when they have positive compatibility with $p_a$ and are ranked by compatibility, shape complementarity, and area ratio. The repair-time cap scales with the movable-item count, total vertex count, and neighborhood scale, subject to the remaining takeover budget. Algorithm~\ref{alg:grounding} summarizes the compilation procedure.

\begin{algorithm}[htbp]
\caption{Failure-grounded neighborhood compilation}
\label{alg:grounding}
\begin{algorithmic}[1]
\Require Layout $L_t$, frontier $\mathcal{F}_t$, pockets $\mathcal{P}_t$, control state $\zeta_t$
\State $\mathcal{A}_t\gets\varnothing$
\For{$f\in\Call{TargetFrontier}{\mathcal{F}_t}$}
    \For{$p\in\Call{DiversePockets}{f,\mathcal{P}_t}$}
        \State $q^\star\gets\Call{ExplanatoryPose}{f,p,L_t}$
        \If{$q^\star$ exists}
            \State $\mathcal{B}^{\mathrm{core}}\gets\mathcal{B}_t(f,q^\star)$
            \For{$m\in\Call{AvailableModes}{\zeta_t}$}
               \State $\mathcal{R}\gets\Call{ExpandRelease}{\mathcal{B}^{\mathrm{core}},m}$
                \State $C\gets\{f\}\cup\Call{PositiveFitAdmit}{f,p}$
                \State $\tau\gets\Call{RepairBudget}{\mathcal{R},C,m,\zeta_t}$
                \If{$\tau$ is executable}
                    \State add $(f,p,\mathcal{R},C,m,\tau)$ to $\mathcal{A}_t$
                \EndIf
            \EndFor
        \EndIf
    \EndFor
\EndFor
\State \Return $\Call{ActionMask}{\Call{BalanceAndCap}{\mathcal{A}_t},\zeta_t}$
\end{algorithmic}
\end{algorithm}

The compiler retains at most $K_A$ candidates. It allocates most slots round-robin across neighborhood modes, covers unseen target--pocket pairs, and backfills remaining slots by the fixed geometric score. The action mask merges candidates inducing the same fixed-item subproblem, suppresses exact repeats, and permits bounded semantic retries. Sustained stagnation enables progressively deeper neighborhoods.

\subsection{Graph Policy}

At decision $t$, $G_t$ encodes each executable repair and its failure geometry. Typed nodes represent targets, pockets, blockers, actions, boundary context, admission context and archived layouts. Object-node attributes encode normalized geometry, failure history, and search context, while action-node attributes encode the fixed geometric score, repair scope, time cap, mode, and remaining-budget ratio. Typed edges encode target--pocket fit, blocker and coupling relations, action membership, and archive dominance.

The actor and critic use independent typed GNN encoders with relation-specific, edge-gated mean aggregation. The actor scores action nodes, while the critic pools non-action nodes to estimate the state value. A raw fixed geometric score $h(a)$ enters the action logits over the eligible set 
\[
z_a=\frac{\ell_\psi(a\mid G_t)+\omega_h h(a)}{\tau_\pi},
\qquad
\pi_\psi=\operatorname{softmax}_{a\in\mathcal{A}_t^{\mathrm{elig}}}(z_a).
\]
Here, $\mathcal{A}_t^{\mathrm{elig}}$ denotes the eligible action set after masking, and $\psi$ denotes the policy parameters. The terms $\ell_\psi(a\mid G_t)$ and $h(a)$ are the learned action logit and deterministic geometric score, respectively, while $\omega_h$ and $\tau_\pi$ are the score weight and policy temperature. Training samples from $\pi_\psi$, whereas inference selects its maximizer.

\subsection{Fixed-Item Repair and Verification}

For action $a$, define
\[
K_a=\mathcal{I}(L_t)\setminus \mathcal{R}_a,
\qquad
I_a=K_a\cup \mathcal{R}_a\cup C_a.
\]
The repair instance contains $I_a$, with items in $K_a$ fixed at their current poses and items in $\mathcal{R}_a\cup C_a$, including $f_a$, optional and movable.

For each repair instance, \solver{} approximates the continuous rotation domain with a finite, geometry-dependent orientation set $\Theta_{\mathrm{exec}}(I_a)$ and searches continuous translations at those orientations within $\tau_a$, returning the feasible candidate with highest $U(L)$.

An internal repair validator enforces fixed poses and checks the merged full layout. The common true-disk verifier then applies the benchmark checks and recomputes $U(L)$. A candidate passing both checks becomes the certified candidate $\widehat L_t$.

If $U(\widehat L_t)>U(B_t)$, we set $L_{t+1}=B_{t+1}=\widehat L_t$. If $U(\widehat L_t)=U(L_t)=U(B_t)$ and the candidate creates a new search context with different action-relevant geometry, we set $L_{t+1}=\widehat L_t$ and retain $B_{t+1}=B_t$ under the bounded plateau rule. If no certified candidate is returned or neither condition holds, we retain $L_{t+1}=L_t$ and $B_{t+1}=B_t$. Thus, $B_{t+1}$ remains feasible and satisfies $U(B_{t+1})\ge U(B_t)$.

\subsection{Learning Objective}

Because each repair decision updates the eligible actions and remaining budget and may change the residual pockets, neighborhood selection is sequential, and the learning objective balances immediate certified improvement, repair cost, and downstream packability. Let $\Delta_t^B=U(B_{t+1})-U(B_t)$. We use the step reward with potential-based shaping \cite{ng1999policy}
\[
r_t=\beta_B\Delta_t^B-\beta_c\widehat c_t+\beta_\Phi\bigl[\gamma\Phi(s_{t+1})-\Phi(s_t)\bigr]+r_t^{\mathrm{evt}},
\]
where $\beta_B$, $\beta_c$, and $\beta_\Phi$ are fixed reward coefficients, $\gamma$ is the discount factor, and $\widehat c_t$ is the normalized repair time. The fixed potential $\Phi$ satisfies $\Phi(s)\in[0,1]$ for every state $s$ and assigns higher values to promising states based on unresolved-item burden, residual placeability, and layout quality. The event term contains clipped degraded-candidate feedback and fixed outcome penalties. At terminal step $H$, it credits improving transitions in proportion to $U(B_H)-U(L_0)$. A terminal correction preserves telescoping over the finite horizon.

We train the masked actor--critic using PPO and GAE \cite{schulman2016gae,schulman2017ppo}. During training, bounded probes compare alternative eligible actions by verifier-certified outcome, and the PPO loss includes a low-weight preference term favoring the better action. Only the policy-selected action advances the trajectory, and training uses fixed, certified initial layouts. Complete takeover pseudocode and implementation details are provided in the supplementary material.

\section{Experiments}
\label{sec:experiments}

\begin{table*}[t!]
\centering
{\small
\setlength{\tabcolsep}{3.0pt}
\renewcommand{\arraystretch}{1.08}
\begin{tabular*}{\textwidth}{@{\extracolsep{\fill}}lcccccc@{}}
\toprule
Source ($N$)
& 2DNesting
& Shadoks-MS
& PackingSolver
& \method{} ($\mu\!\pm\!\sigma$)
& Gain [95\% CI]
& W/T/L \\
\midrule
ESICUP (90)
& 64.98
& 72.97
& 78.59
& $\mathbf{79.55}\boldsymbol{\pm}\mathbf{0.08}$
& $+0.96\,[0.64,1.31]$
& $62/1/27$ \\

Synthetic (90)
& 62.00
& 72.47
& 79.61
& $\mathbf{80.64}\boldsymbol{\pm}\mathbf{0.08}$
& $+1.03\,[0.70,1.39]$
& $64/0/26$ \\

Gardeyn (54)
& 60.47
& 72.17
& 76.93
& $\mathbf{77.53}\boldsymbol{\pm}\mathbf{0.11}$
& $+0.60\,[0.34,0.86]$
& $36/2/16$ \\

Main Avg. (234)
& 62.49
& 72.54
& 78.37
& $\mathbf{79.24}\boldsymbol{\pm}\mathbf{0.08}$
& $+0.87\,[0.69,1.05]$
& $162/3/69$ \\
\bottomrule
\end{tabular*}
}
\caption{Main results under matched 650-s budgets. Entries report verifier-certified utilization (\%). \method{} is reported as the mean and sample standard deviation over three training seeds, and Main Avg. equally weights the three sources.}
\label{tab:main}
\end{table*}

\subsection{Experimental Setup}

\paragraph{Data and training.}
We evaluate all methods on the held-out test partitions in Tab.~\ref{tab:benchmark-splits}. For each non-CAD source, we train three policies using seeds 42, 3407, and 7301, with checkpoints selected on a fixed source-specific validation set and frozen before testing. Results are averaged per instance across seeds before source aggregation, and Main Avg. equally weights ESICUP, Synthetic, and Gardeyn. All nine policies are evaluated zero-shot on CAD, with results grouped by training source. Training, validation, and tuning use only the three non-CAD sources.

\paragraph{Baselines.}
Reproducible competitive irregular-packing implementations remain scarce \citep{gardeyn2025sparrow}. We compare against \textsc{PackingSolver} (PS) \citep{packingsolver}, \textsc{2DNesting} \citep{twodnesting}, and \textsc{Shadoks-MS} \citep{dafonseca2024shadoks}. All three baselines are deterministic under our configurations and evaluated once per instance. PS is used without algorithmic modification, while \textsc{2DNesting} and \textsc{Shadoks-MS} are adapted to the circular-container knapsack setting. All methods solve identical instances under matched wall-clock budgets and search orientations over the full $[0^\circ,360^\circ)$ range using their own placement and orientation-search strategies.

\paragraph{Protocol and metrics.}
Under the 650-s budget, standalone methods use the full budget, while \method{} uses 350 s for PS initialization and 300 s for takeover. Validation curves showed that PS plateaued by 350 s, so this split was fixed before testing. Under the 1200-s budget, PS uses the full budget, while \method{} uses 600 s each for initialization and takeover. Failure-geometry extraction, action and graph construction, policy inference, repair, in-loop verification, rollback, and controller bookkeeping count toward the takeover budget.

All methods are evaluated on the same Intel Xeon Platinum 8352V node with 125 GiB RAM and one CPU thread per instance, while training uses one NVIDIA RTX A5000 GPU. We report verifier-certified mean utilization, paired gain in percentage points, and instance-level W/T/L against PS. After averaging over training seeds, 95\% confidence intervals are computed from 200,000 paired cluster-bootstrap resamples of base candidate-item sets within each source, with all load variants of each sampled set retained together. Sources are resampled separately before computing Main Avg. Selector comparisons use two-sided paired sign-flip tests at the base-set level with Holm correction.

\subsection{Main Comparison under Matched Budgets}

\begin{figure}[htbp]
    \centering
    \includegraphics[width=\columnwidth]{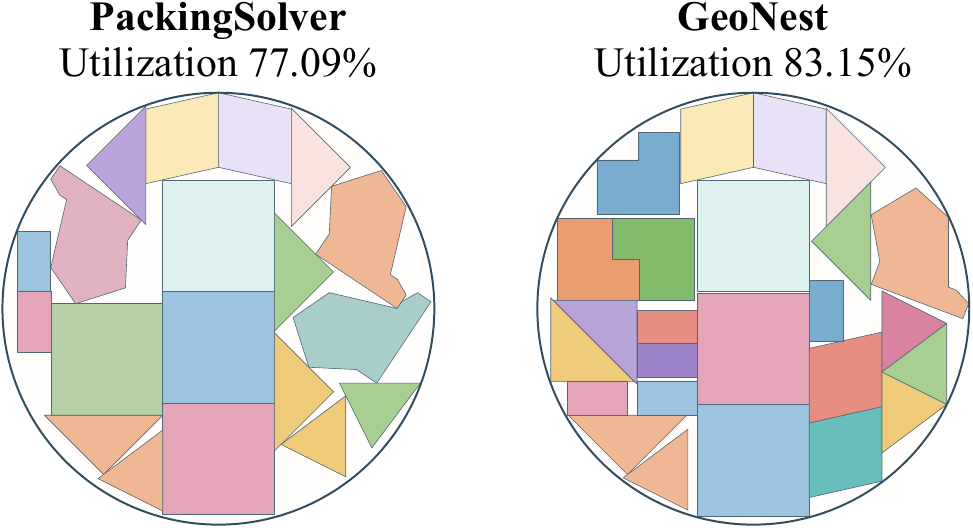}
    \caption{Verifier-certified PackingSolver and \method{} layouts for the same instance under matched 650-s budgets.}
    \label{fig:layout-comparison}
\end{figure}

\paragraph{Matched-budget performance.}
Tab.~\ref{tab:main} shows positive gains across all three sources, and all paired 95\% confidence intervals exclude zero. PackingSolver already reaches a Main Avg. utilization of $78.37\%$. At this utilization level, further improvement increasingly depends on reorganizing constrained residual space rather than filling obvious gaps. These results support a late-stage shift in search strategy after geometric construction has matured, redirecting computation from continued generic improvement to failure-conditioned repair, as illustrated in Fig.~\ref{fig:layout-comparison}.

\paragraph{Budget robustness and efficiency.}
The matched 1200-s comparison shows positive gains across all three sources. Despite using 45.8\% less time, the 650-s \method{} configuration exceeds PackingSolver at 1200 s by 0.46 pp. The matched-budget gain at 1200 s retains 88.5\% of the 650-s gain, showing that the benefit persists as both methods receive more time. This cross-budget reversal shows that targeted repair is more effective than nearly doubling generic search time.

\begin{table}[htbp]
\centering
{\small
\setlength{\tabcolsep}{1.0pt}
\renewcommand{\arraystretch}{1.06}
\begin{tabular*}{\columnwidth}{@{\extracolsep{\fill}}lccc@{}}
\toprule
\multicolumn{4}{l}{\textit{(a) Matched 1200-s budget}} \\
Source
& PS
& \method{} ($\mu\!\pm\!\sigma$)
& Gain [95\% CI] \\
\midrule
ESICUP
& 78.81
& $\mathbf{79.65}\boldsymbol{\pm}\mathbf{0.10}$
& $+0.84\,[0.59,1.14]$ \\
Synthetic
& 80.06
& $\mathbf{80.96}\boldsymbol{\pm}\mathbf{0.13}$
& $+0.90\,[0.68,1.14]$ \\
Gardeyn
& 77.46
& $\mathbf{78.03}\boldsymbol{\pm}\mathbf{0.13}$
& $+0.57\,[0.31,0.82]$ \\
Main Avg.
& 78.78
& $\mathbf{79.55}\boldsymbol{\pm}\mathbf{0.11}$
& $+0.77\,[0.63,0.92]$ \\

\midrule
\multicolumn{4}{l}{\textit{(b) Cross-budget efficiency}} \\
Source
& PS
& \method{} ($\mu\!\pm\!\sigma$)
& Gain [95\% CI] \\
\midrule
ESICUP
& 78.81
& $\mathbf{79.55}\boldsymbol{\pm}\mathbf{0.08}$
& $+0.74\,[0.50,1.01]$ \\
Synthetic
& 80.06
& $\mathbf{80.64}\boldsymbol{\pm}\mathbf{0.08}$
& $+0.58\,[0.32,0.85]$ \\
Gardeyn
& 77.46
& $\mathbf{77.53}\boldsymbol{\pm}\mathbf{0.11}$
& $+0.07\,[-0.17,0.30]$ \\
Main Avg.
& 78.78
& $\mathbf{79.24}\boldsymbol{\pm}\mathbf{0.08}$
& $+0.46\,[0.32,0.61]$ \\
\bottomrule
\end{tabular*}
}
\caption{Long-budget results with verifier-certified utilization (\%). Panel (a) reports matched 1200-s comparisons. Panel (b) compares 1200-s PackingSolver with the 650-s \method{} configuration.}
\label{tab:budget}
\end{table}

Gardeyn is the only source with an inconclusive cross-budget comparison. Its more concave and intricate polygons induce complex pockets and blocker relations. The positive matched-budget gain and weaker cross-budget result are consistent with such geometry requiring a longer repair horizon. 

\subsection{Ablation and Generalization}

\paragraph{Component analysis.}
All variants reuse the same 350-s PackingSolver layout and receive 300 s. Generic-LNS is failure-blind. FA-LNS-Random, FA-LNS-Score, FA-LNS-MLP, and \method{} use the same state-dependent failure-aware action-generation procedure and repair executor. They differ only in action selection, using random choice, a fixed score, an MLP, or the relation-aware GNN.

\begin{table}[htbp]
\centering
{\small
\renewcommand{\arraystretch}{1.06}

\setlength{\tabcolsep}{1.2pt}
\begin{tabular*}{\columnwidth}
{@{\extracolsep{\fill}}lccccc@{}}
\toprule
\multicolumn{6}{l}{\textit{(a) Component analysis}} \\
Variant & E & S & G & Main & $\Delta$ \\
\midrule
PackingSolver
& 78.59 & 79.61 & 76.93 & 78.37 & -- \\

Generic-LNS
& 79.16 & 79.89 & 77.17 & 78.74 & $+0.37$ \\

FA-LNS-Random
& 79.32 & 80.43 & 77.34 & 79.03 & $+0.66$ \\

FA-LNS-Score
& 79.41 & 80.53 & 77.34 & 79.10 & $+0.73$ \\

FA-LNS-MLP
& 79.34 & 80.58 & 77.49 & 79.14 & $+0.77$ \\

\method{}
& \textbf{79.55}
& \textbf{80.64}
& \textbf{77.53}
& \textbf{79.24}
& $\mathbf{+0.87}$ \\
\end{tabular*}\par

\vspace{1.5pt}
\noindent\rule{\columnwidth}{0.4pt}\par
\vspace{1.5pt}

\setlength{\tabcolsep}{2.0pt}
\begin{tabular*}{\columnwidth}
{@{\extracolsep{\fill}}lccc@{}}
\multicolumn{4}{l}{\textit{(b) Direct selector and architecture comparisons}} \\
Comparison
& Gain
& 95\% CI
& $p_{\mathrm{Holm}}$ \\
\midrule
\method{} $-$ Random
& $+0.21$
& $[0.13,0.29]$
& $1.50{\times}10^{-5}$ \\

\method{} $-$ Score
& $+0.14$
& $[0.04,0.26]$
& $1.95{\times}10^{-2}$ \\

\method{} $-$ MLP
& $+0.10$
& $[0.03,0.18]$
& $1.95{\times}10^{-2}$ \\
\bottomrule
\end{tabular*}
}
\caption{Component analysis. E, S, and G denote ESICUP, Synthetic, and Gardeyn. Panel (a) reports verifier-certified utilization (\%), with $\Delta$ denoting the equal-source Main improvement over PackingSolver. Panel (b) compares selectors under the same failure-aware action-generation procedure. \method{}, FA-LNS-Random, and FA-LNS-MLP are averaged over three seeds.}
\label{tab:ablation}
\end{table}

\paragraph{Failure-aware actions and learned selection.}
Failure-aware actions account for most of the gain beyond Generic-LNS. Because all selectors use the same action-generation procedure and repair executor, Tab.~\ref{tab:ablation}(b) isolates the value of action selection. \method{} significantly outperforms Random, Score, and MLP, showing that relation-aware modeling of the target--pocket--blocker structure adds value beyond heuristic and non-graph learned ranking. A one-step audit of 18 states, for which all candidate actions were evaluated, found that the GNN selected the best immediate repair in 61.1\% of cases and achieved a higher immediate gain than the mean over candidate actions, linking the end-to-end improvement to action-level discrimination.

\paragraph{Scale and load.}
Tab.~\ref{tab:robustness} shows opposite trends for instance size and packing pressure. Gain decreases with $n$ but increases with $\lambda$, indicating that \method{} responds more strongly to spatial competition than to instance size alone. With larger item sets, a fixed repair budget is spread across more candidate neighborhoods, whereas higher load magnifies the impact of blockers and residual pockets.

\begin{table}[htbp]
\centering
{\small
\renewcommand{\arraystretch}{1.05}

\setlength{\tabcolsep}{1.5pt}
\begin{tabular*}{\columnwidth}
{@{\extracolsep{\fill}}lccc@{}}
\toprule
\multicolumn{4}{l}{\textit{(a) Main-set marginal trends}} \\
Factor & Low & Medium & High \\
\midrule
$n$
& $48\,(+1.00)$
& $64\,(+0.93)$
& $96\,(+0.67)$ \\
$\lambda$
& $1.2\,(+0.71)$
& $1.5\,(+0.79)$
& $2.0\,(+1.10)$ \\
\end{tabular*}\par

\vspace{1.5pt}
\noindent\rule{\columnwidth}{0.4pt}\par
\vspace{1.5pt}

\setlength{\tabcolsep}{0.5pt}
\begin{tabular*}{\columnwidth}
{@{\extracolsep{\fill}}lcccc@{}}
\multicolumn{5}{l}{\textit{(b) Held-out shifts}} \\
Shift ($N$)
& PackingSolver
& \method{} ($\mu\!\pm\!\sigma$)
& Gain
& W/T/L \\
\midrule
Size (84)
& 80.70
& $\mathbf{80.92}\boldsymbol{\pm}\mathbf{0.07}$
& $+0.22$
& $63/10/11$ \\
Stress (25)
& 77.86
& $\mathbf{78.56}\boldsymbol{\pm}\mathbf{0.08}$
& $+0.70$
& $18/2/5$ \\
\bottomrule
\end{tabular*}
}
\caption{Scale and load robustness under the 650-s budget. Panel (a) reports gains averaged over three seeds and equally across sources, with each cell showing the factor value and gain in pp. Panel (b) reports equal-source utilization (\%) and gain, together with instance-level W/T/L, on the held-out Size and Stress shifts.}
\label{tab:robustness}
\end{table}

The held-out shifts reinforce this interpretation. Stress yields 3.2 times the mean gain of Size despite similar win rates. This pattern suggests that packing pressure affects the magnitude of improvement more than the frequency of wins, consistent with future-feasibility loss driven by residual-space competition.

\paragraph{Cross-source and industrial transfer.}

All six off-diagonal gains are positive, and their mean is 91.7\% of the mean source-matched gain. Policies trained on each of the three sources also produce similar positive CAD gains with no instance-level losses. The identical W/T/L counts across training sources further indicate consistent transfer to CAD. Together, these results show that relational failure cues transfer across contour distributions without requiring close contour matching.

\begin{table}[htbp]
\centering
{\small
\renewcommand{\arraystretch}{1.06}

\setlength{\tabcolsep}{3.0pt}
\begin{tabular*}{\columnwidth}
{@{\extracolsep{\fill}}lccc@{}}
\toprule
\multicolumn{4}{l}{\textit{(a) Cross-source policy transfer}} \\
Train $\backslash$ Test
& ESICUP
& Synthetic
& Gardeyn \\
\midrule
ESICUP
& $+0.96$
& $+0.93$
& $+0.63$ \\
Synthetic
& $+0.80$
& $+1.03$
& $+0.61$ \\
Gardeyn
& $+0.89$
& $+0.89$
& $+0.60$ \\
\end{tabular*}\par

\vspace{1.5pt}
\noindent\rule{\columnwidth}{0.4pt}\par
\vspace{1.5pt}

\setlength{\tabcolsep}{0.5pt}
\renewcommand{\arraystretch}{1.05}
\noindent
\begin{tabular*}{\columnwidth}
{@{\extracolsep{\fill}}lrrrr@{}}
\multicolumn{5}{l}{\textit{(b) Held-out industrial CAD transfer}} \\
Train
& \multicolumn{1}{c}{PS}
& \multicolumn{1}{c}{\method{}}
& \multicolumn{1}{c}{Gain [95\% CI]}
& \multicolumn{1}{c}{W/T/L} \\
\midrule
ESICUP
& 79.78
& \textbf{80.39}
& $+0.61\,[0.26,1.01]$
& $11/3/0$ \\
Synthetic
& 79.78
& \textbf{80.46}
& $+0.68\,[0.32,1.19]$
& $11/3/0$ \\
Gardeyn
& 79.78
& \textbf{80.39}
& $+0.61\,[0.26,1.08]$
& $11/3/0$ \\
\bottomrule
\end{tabular*}
}
\caption{Generalization under the 650-s budget. Panel (a) reports seed-averaged gains over PS, with off-diagonal entries denoting zero-shot transfer. Panel (b) reports verifier-certified results on 14 held-out industrial CAD instances.}
\label{tab:generalization}
\end{table}

\section{Conclusion}

We formalized 2D-CIKP and introduced \bench{} for reproducible, verifier-backed evaluation. 
\method{} treats unpacked items as evidence of residual pockets, explanatory poses, and blockers, then uses a relation-aware graph policy to select which fixed-item repair neighborhood to execute within the budget. 
Ablations show failure grounding drives most gains beyond Generic-LNS, while learned selection significantly outperforms random, fixed-score, and MLP selectors. 
Starting from PackingSolver incumbents, \method{} consistently yields substantial utilization improvements under matched budgets. 
Notably, our approach achieves cross-budget efficiency, where shorter-budget configurations outperform the standalone solver under extended limits. 
Gains strengthen under higher packing pressure, and the learned policy
transfers across unseen contour sources and held-out industrial CAD
geometries.
Together, the results show that late-stage packing is limited less by insufficient search than by failure diagnosis and repair allocation, positioning learning to direct expensive geometric computation rather than replacing it with pure neural approximations. 
The fixed-item repair contract enables this separation, allowing stronger compatible solvers to raise the backend ceiling while preserving the same failure-driven decision layer.

\bibliography{aaai2027}

\end{document}